\pdfoutput=1

\documentclass[11pt]{article}

\usepackage[]{acl} 

\usepackage{times}
\usepackage{latexsym}

\usepackage[T1]{fontenc}

\usepackage[utf8]{inputenc}

\usepackage{microtype}

\usepackage{nicefrac}
\usepackage{xfrac}
\usepackage{tabularx, graphicx} 
\usepackage{booktabs}
\usepackage{subcaption}
\usepackage{soul}
\usepackage{amssymb}
\usepackage{pifont}
\usepackage{longtable,array}
\usepackage{lipsum} 
\usepackage{hhline}
\usepackage[bottom]{footmisc}

\title{Approximating Online Human Evaluation of Social Chatbots with Prompting}

\author{Ekaterina Svikhnushina \and Pearl Pu \\
        School of Computer and Communication Sciences \\ EPFL, Lausanne, Switzerland \\
        \texttt{\{ekaterina.svikhnushina,pearl.pu\}@epfl.ch}}

\begin{document}

\maketitle
\begin{abstract}
As conversational models become increasingly available to the general public, users are engaging with this technology in social interactions. Such unprecedented interaction experiences may pose considerable social and psychological risks to the users unless the technology is properly controlled. This highlights the need for scalable and robust evaluation metrics for conversational chatbots. Existing evaluation metrics aim to automate offline user evaluation and approximate human judgment of pre-curated dialogs. However, they are limited in their ability to capture subjective perceptions of users who actually interact with the bots and might not generalize to real-world settings. To address this limitation, we propose an approach to approximate online human evaluation leveraging large language models (LLMs) from the GPT family. We introduce a new Dialog system Evaluation framework based on Prompting (DEP), which enables a fully automatic evaluation pipeline that replicates live user studies and achieves an impressive correlation with human judgment (up to Pearson $r=0.95$ on a system level). The DEP approach involves collecting synthetic chat logs of evaluated bots with an LLM in the other-play setting, where the LLM is carefully conditioned to follow a specific scenario. We further explore different prompting approaches to produce evaluation scores with the same LLM. The best-performing prompts, which contain few-shot demonstrations and instructions, show outstanding performance on the tested dataset and demonstrate the ability to generalize to other dialog corpora.
\end{abstract}

\section{Introduction}

The recent arrival of conversational AI, marked by the public release of ChatGPT from OpenAI,\footnote{\url{https://openai.com/blog/chatgpt}} initiated unprecedented user engagement with conversational chatbots in a real-world setting. With the impressive naturalness of machines' responses, users are going beyond traditional transactional exchanges and start exploring more social interaction scenarios with increasing curiosity \cite{statista2023}. In such situations, users might be subject to social and psychological harms if dialog systems fail to follow commonsense social rules \cite{Svikhnushina2022PEACE, kim-etal-2022-prosocialdialog}. Several instances of alarming social behavior of this technology have already been discussed in the media \cite{nyt2023, sa2022, openletter2023}. In this context, developing meaningful and robust evaluation metrics for these systems has become particularly urgent to ensure that the models are safe and acting in the best interest of the users before their release.

\begin{figure*}[!t]
    \centering
    \includegraphics[width=0.84\linewidth]{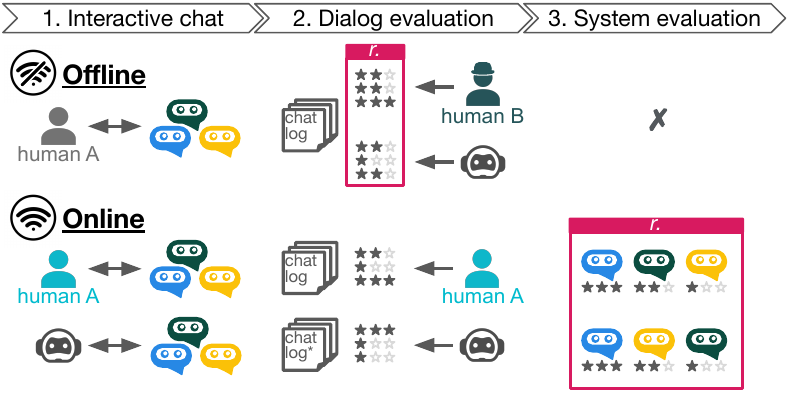}
    \caption{Offline and online dialog evaluation with the corresponding processes. In the first step, dialog logs are curated. In the second step, each dialog log is assigned a dialog-level score, either by a third-party judge (offline) or by the same conversational partner (online). In the third step, the system ranking is obtained by aggregating the dialog scores of each chatbot. Grey bot icons indicate steps that are intended to be approximated by means of automatic evaluation.
    Pink boxes mark the steps in the process where the correlation ($r.$) with the ground truth human judgment is computed to validate the automatic evaluation metric during its development process.}
    \label{fig:eval_types}
\end{figure*}

Initially, human evaluation was considered a de facto standard for evaluating dialog systems \cite{li2019acute}. As running human evaluation is time- and resource-consuming, a number of automatic evaluation metrics for dialog systems have been proposed \cite{mehri2022report, yeh-etal-2021-comprehensive}. The majority of these approaches aim to automate the \textit{offline} user evaluation. In this setting, dialog evaluation is performed by a human judge who is distinct from the one conversing with the bot (Figure \ref{fig:eval_types}, offline). The metrics proposed for this case approximate the evaluation scores provided by this third-party human judge for the pre-produced dialogs \cite[\textit{e.g.}][]{mehri-eskenazi-2020-unsupervised, ghazarian-etal-2022-wrong}. Despite its popularity, offline user evaluation is limited in its ability to capture subjective perceptions of users who actually interacted with the bots \cite{jannach2022evaluating, lee2022evaluating, NEURIPS2019_fc981212}. This limitation of relying on second-hand evaluation can be illustrated by an analogy from the realm of restaurant critique when one tries to evaluate a restaurant solely by reading consumer reviews but having never actually eaten there. Conducting \textit{online} user evaluation, where the same individual interacts with the bot and assesses its performance, is more likely to produce accurate and precise evaluations of the chatbot's performance. Moreover, this method offers better predictive capabilities for the system use ``in the wild'' \cite{Beel2015}. However, by far, efforts towards approximating online user evaluation have been limited.

To address this gap, we propose a novel automatic \textbf{D}ialog system \textbf{E}valuation framework based on \textbf{P}rompting, DEP. Our framework automates the whole pipeline of dialog system evaluation in an interactive setting, replicating live user studies. As the first step towards this goal, we leverage a large language model (LLM) from the GPT-family models to collect synthetic chat logs of evaluated bots with the LLM. Second, we prompt the same LLM to produce the resulting evaluation scores for generated chat logs and, finally, rank the chatbots based on their overall performance (Figure \ref{fig:eval_types}, online).

While using bot-play is not a new idea per se, we emphasize the importance of carefully choosing a dialog partner for the evaluated chatbots specifically for social conversational contexts where the roles of two interlocutors can differ significantly. For example, it was shown that the emotion/intent distributions in conversations between an emotional speaker and an empathetic listener are very different for the two dialog partners \cite{welivita-pu-2020-taxonomy}. To account for it, in the first step of our framework, we propose prompting LLMs to play a particular social role over the course of the interaction with the chatbots to be evaluated. For the second step, we draw inspiration from the fact that LLMs demonstrate solid performance improvement when their generation process is augmented with instructions \cite{kim-etal-2022-prosocialdialog}. We demonstrate that prompting the model with appropriate instructions that explain how fine-grained evaluation dimensions relate to the overall dialog score leads to substantial performance improvement, reaching up to $r=0.95$ Pearson correlation with the human judgment on a system level.

Overall, our contributions include the following. 1) We describe an end-to-end prompting-based evaluation framework for dialog systems, specifically targeting social interaction scenarios (Section \ref{sec:proposed_method}). 2) Our experiments showcase the effectiveness of prompting for assigning a desired social role to LLMs and, thus, collecting machine-generated dialogs that better approximate real interpersonal communication (Section \ref{sec:llm-to-bot}). 3) We consider different prompt designs and conclude that including demonstrations together with instructions results in the best performance (Sections \ref{sec:prompted_eval}, \ref{sec:prompted_eval_ext}).

\section{Related Work}

\subsection{Automatic Evaluation of Chatbots}
Automatic dialog evaluation has been a long-standing research topic for practitioners. Initial works focused on evaluating chatbots' responses against a ground-truth reference \cite{papineni2002bleu, tao2018ruber}. Following works moved on to exploring reference-free evaluation metrics as the referenced evaluation was shown to be ineffective due to a wide range of acceptable responses for a single context \cite{liu-etal-2016-evaluate}, implying that comparing with a single reference is limited. Reference-free metrics usually operate either on the utterance or the dialog level. For the utterance level, practitioners have explored ways to evaluate response appropriateness for the preceding context \cite{Lan2020, pang-etal-2020-towards} or predict the qualities of the follow-up response as a proxy for the quality of the preceding dialog \cite{ghazarian-etal-2022-wrong, ghazarian2020predictive, mehri-eskenazi-2020-unsupervised}. For the dialog level, a number of diverse approaches have been proposed, ranging from aggregating several fine-grained utterance-level evaluations \cite{Zhang2021}, to designing training objectives to model the information flow across dialogue utterances \cite{li-etal-2021-conversations}, employing graph representations to capture dialog dynamics \cite{huang-etal-2020-grade, zhang-etal-2021-dynaeval}, and using semantic-level manipulations to teach the evaluation model to distinguish coherent and incoherent dialogs \cite{ghazarian-etal-2022-deam}.

The works above largely target the offline evaluation setting. Some scholars have also started exploring different ways of approximating online user evaluation. \citet{deriu-etal-2020-spot} proposed a partially automated framework where human judges rank chatbots regarding their ability to mimic conversational behavior using interactively collected bot-to-bot conversations, which relies on survival analysis. \citet{sato-etal-2022-bipartite} proposed a particular bipartite-play approach for collecting bot-to-bot conversations to provide a fairer comparison setting for evaluated chatbots. These papers consider methodologies for organizing bot-to-bot conversation sessions, but they are not concerned with the way how these bot-to-bot conversations unfold. In our work, we explore the use of bot-to-bot conversations to model a desired social behavior.

\subsection{Prompting}
Prompt-based learning paradigm \cite{Liu2023} received significant attention after \citet{brown2020language} demonstrated how GPT-3, a large foundation model, can well handle a wide range of tasks without the need for fine-tuning, relying only on natural-language prompts and task demonstrations as context. Prompt-based model performance depends on the design of the provided prompt. Prompt engineering efforts explore approaches for designing prompts, which vary in the shape of prompts (cloze or prefix), human effort required for writing prompts (manual or automatic), and number of demonstrations provided to the model in the prompt (zero-shot or few-shot) \cite{Liu2023}. 

Prompt-based learning applied to recently created LLMs has been reported to achieve outstanding results on a variety of tasks and benchmarks, including classification, reasoning, coding, translation, and many others \cite[\textit{e.g.}][]{wei2022emergent, chowdhery2022palm, chung2022scaling}. However, exploring prompting for the evaluation of dialog systems has not been widely investigated. We are only aware of one more simultaneous and independent effort in this direction. \citet{huynh2023understanding} studied how different LLM parameters (type, size, training data) may influence the dialog evaluation, focusing on utterance- and dialog-level evaluation in the offline evaluation setting.
Our work focuses on how prompting can be used to capture a holistic evaluation of dialog systems in online social settings, relying on freshly generated dialogs.

\section{Proposed Method: DEP} \label{sec:proposed_method}
We introduce our DEP framework, which consists of two consecutive steps. First, it requires collecting interactive chat logs between the LLM and evaluated chatbots, which we denote as LLM-to-bot play. Second, the LLM is prompted to generate scores for these chat logs. The generated scores are further aggregated to produce a final ranking of the systems. We describe each of the steps below.

\subsection{Prompted LLM-to-Bot Play}
In social settings, two partners may play considerably different roles in a dialog, thus establishing very distinct conversational behaviors. Examples include conversations between a student and a teacher, an emotional speaker and an empathetic listener, or even between two interlocutors with different personas. Chatbots are usually built to perform well in one of these roles (e.g., empathetic listener), but not necessarily the other. Therefore, collecting synthesized dialogs via self-play of the chatbot with itself (or a similar competing model) might fail to represent a realistic discourse flow due to the differences in the intents produced by speakers and listeners in dialogs.

To address this consideration and render the synthesized dialogs that better approximate real social interactions, we propose leveraging LLMs' ability to produce responses on behalf of an assigned character \cite{thoppilan2022lamda}. Specifically, we suggest letting the evaluated chatbots converse with an LLM prompted to play a particular social role. Figure \ref{fig:LLM-to-bot} demonstrates how to structure the prompt to produce each next output of the LLM in an interactive manner. Meanwhile, responses from the evaluated chatbots are computed by passing the accumulated dialog history to these chatbots as input context. The process can be repeated for multiple dialog turns. The length of the exchange may depend on the extent of details provided to prompt the LLM. The more specific the prompt is, the faster the evaluated chatbot can demonstrate its performance in the social situation of interest. On the contrary, more generic conversation starters require more dialog turns to reveal the targeted social behavior.

\begin{table}[]
\centering

\begin{tabular}{|p{.15\linewidth} p{.7\linewidth}|}
\hline
\multicolumn{2}{|p{.9\linewidth}|}{I am a Speaker \textit{<in an assigned social situation>}. I am sharing \textit{<my thoughts>} with a Listener in a dialog.} \\ 
Speaker:  & \textit{<LLM's input \#1>}                    \\
Listener: & \textit{<Bot's response \#1>}                 \\
Speaker:  &                                               \\ \hline
\end{tabular}
\captionof{figure}{Prompt template to condition a LLM to play an assigned social role while interacting with an evaluated chatbot.} \label{fig:LLM-to-bot}
\end{table}

\subsection{Prompted Evaluation}
Once dialog logs are synthesized, we propose using prompting to produce evaluation scores for each dialog. Prompts can be constructed in several ways. We investigate zero-shot and few-shot settings, either with or without instructions, in our experiments (Section \ref{sec:results}). Many available foundation LLMs are accessible through APIs and only output text completions without corresponding log probabilities. Therefore, regardless of the type of prompt that we use, to generate a score for each dialog, we obtain a textual form of the score from the LLM completion and then use a verbalizer function to map it to a numerical value, getting inspiration from \cite{schick-schutze-2021-exploiting}. Formally, given a dialog log $d$, we construct a prompt $P(d)$ that takes $d$ as input and outputs a prompt that contains exactly one mask token as a placeholder for the dialog score. Let $y$ be a predicted token for $P(d)$. We then define a verbalizer as an injective function $v$ that maps each score in textual form to a numerical value. Thus, $v(y)$ produces a numerical score for a single dialog. The final rating of a given dialog system is obtained by averaging the corresponding dialog scores of that system. For fair evaluation, the number of dialogs collected for each evaluated chatbot should be identical.

\section{Results} \label{sec:results}
For all reported experiments, we used the most capable version of the InstructGPT model ({\fontfamily{qcr}\selectfont text-davinci-003}) available at the moment of initiation of our experiments in early Q1 2023. We used this model as it was easily accessible through OpenAI API\footnote{\url{https://openai.com/blog/openai-api}} and was expected to have superior performance for social scenarios as it was trained based on human feedback, which captures subjective human judgment of interactive outputs \cite{ouyang2022training}.

Following previous works that considered system-level evaluation \cite{lowe-etal-2017-towards, NEURIPS2019_fc981212}, we report Pearson correlation for our experiments, unless specified otherwise. We also opted for this type of correlation coefficient as it performed better for capturing whether the automated metric succeeds in preserving the gap in scores for the best- and least-performing chatbots, the information which gets lost with rank correlation.

We start by demonstrating the application of our evaluation framework to empathetic dialog systems as in these interactive scenarios two conversational partners have clearly distinct social roles: an emotional speaker and an empathetic listener. Further, we consider the generalizing ability of the framework to other social domains.

\subsection{Evaluation of Empathetic Chatbots}
Below, we first describe the dataset used for the experiment. Then, we consider the ability of prompted LLM to effectively replicate social discourse patterns over multi-turn interactions with the chatbots that serve as eventual evaluation targets. Finally, we explore several types of prompts applied to synthesized LLM-to-bots dialogs to evaluate how well they can approximate human judgment on a system level.

\subsubsection{Dataset and Evaluated Chatbots}

\begin{table*}[h!]
\centering
\resizebox{\textwidth}{!}{%
\begin{tabular}{cc|cc|cc}
\toprule
\multicolumn{2}{c|}{Turn 2}                                                                                                                         & \multicolumn{2}{c|}{Turn 4}                                                                                                                          & \multicolumn{2}{c}{Turn 6}                                                                                                                           \\ \midrule
human $\leftrightarrow$ \textbf{bot}                                                                    & LLM $\leftrightarrow$ \textbf{bot}                                                                      & human $\leftrightarrow$ \textbf{bot}                                                                    & LLM $\leftrightarrow$ \textbf{bot}                                                                       & human $\leftrightarrow$ \textbf{bot}                                                                    & LLM $\leftrightarrow$ \textbf{bot}                                                                       \\ \midrule
\begin{tabular}[c]{@{}c@{}}questioning\\      \small{2033; 53.0\%}\end{tabular}    & \begin{tabular}[c]{@{}c@{}}questioning\\      \small{2030; 52.9\%}\end{tabular}  & \begin{tabular}[c]{@{}c@{}}questioning\\      \small{1336; 34.8\%}\end{tabular}  & \begin{tabular}[c]{@{}c@{}}acknowledging\\      \small{1148; 29.9\%}\end{tabular} & \begin{tabular}[c]{@{}c@{}}questioning\\      \small{1062; 27.7\%}\end{tabular}  & \begin{tabular}[c]{@{}c@{}}acknowledging\\      \small{1261; 32.8\%}\end{tabular} \\
\begin{tabular}[c]{@{}c@{}}sympathizing\\      \small{716; 18.7\%}\end{tabular}  & \begin{tabular}[c]{@{}c@{}}sympathizing\\      \small{710; 18.5\%}\end{tabular}  & \begin{tabular}[c]{@{}c@{}}acknowledging\\      \small{770; 20.1\%}\end{tabular} & \begin{tabular}[c]{@{}c@{}}questioning\\      \small{916; 23.9\%}\end{tabular}    & \begin{tabular}[c]{@{}c@{}}acknowledging\\      \small{881; 22.9\%}\end{tabular} & \begin{tabular}[c]{@{}c@{}}questioning\\      \small{550; 14.3\%}\end{tabular}    \\
\begin{tabular}[c]{@{}c@{}}acknowledging\\      \small{528; 13.8\%}\end{tabular} & \begin{tabular}[c]{@{}c@{}}acknowledging\\      \small{534; 13.9\%}\end{tabular} & \begin{tabular}[c]{@{}c@{}}sympathizing\\      \small{554; 14.4\%}\end{tabular}  & \begin{tabular}[c]{@{}c@{}}sympathizing\\      \small{527; 13.7\%}\end{tabular}   & \begin{tabular}[c]{@{}c@{}}sympathizing\\      \small{494; 12.9\%}\end{tabular}  & \begin{tabular}[c]{@{}c@{}}encouraging\\      \small{464; 12.1\%}\end{tabular}    \\
\begin{tabular}[c]{@{}c@{}}encouraging\\      \small{168; 4.4\%}\end{tabular}    & \begin{tabular}[c]{@{}c@{}}encouraging\\      \small{164; 4.3\%}\end{tabular}    & \begin{tabular}[c]{@{}c@{}}encouraging\\      \small{266; 6.9\%}\end{tabular}    & \begin{tabular}[c]{@{}c@{}}encouraging\\      \small{354; 9.2\%}\end{tabular}     & \begin{tabular}[c]{@{}c@{}}encouraging\\      \small{376; 9.8\%}\end{tabular}    & \begin{tabular}[c]{@{}c@{}}sympathizing\\      \small{448; 11.7\%}\end{tabular}   \\
\begin{tabular}[c]{@{}c@{}}consoling\\      \small{126; 3.3\%}\end{tabular}      & \begin{tabular}[c]{@{}c@{}}consoling\\      \small{154; 4.0\%}\end{tabular}        & \begin{tabular}[c]{@{}c@{}}neutral\\      \small{228; 5.9\%}\end{tabular}        & \begin{tabular}[c]{@{}c@{}}consoling\\      \small{244; 6.4\%}\end{tabular}       & \begin{tabular}[c]{@{}c@{}}wishing\\      \small{226; 5.9\%}\end{tabular}        & \begin{tabular}[c]{@{}c@{}}wishing\\      \small{338; 8.8\%}\end{tabular}         \\
\begin{tabular}[c]{@{}c@{}}neutral\\      \small{122; 3.2\%}\end{tabular}        & \begin{tabular}[c]{@{}c@{}}neutral\\      \small{97; 2.5\%}\end{tabular}         & \begin{tabular}[c]{@{}c@{}}consoling\\      \small{206; 5.4\%}\end{tabular}      & \begin{tabular}[c]{@{}c@{}}neutral\\      \small{214; 5.6\%}\end{tabular}         & \begin{tabular}[c]{@{}c@{}}neutral\\      \small{192; 5.0\%}\end{tabular}          & \begin{tabular}[c]{@{}c@{}}agreeing\\      \small{250; 6.5\%}\end{tabular}        \\
\begin{tabular}[c]{@{}c@{}}agreeing\\      \small{62; 1.6\%}\end{tabular}        & \begin{tabular}[c]{@{}c@{}}agreeing\\      \small{64; 1.7\%}\end{tabular}        & \begin{tabular}[c]{@{}c@{}}agreeing\\      \small{127; 3.3\%}\end{tabular}       & \begin{tabular}[c]{@{}c@{}}agreeing\\      \small{206; 5.4\%}\end{tabular}        & \begin{tabular}[c]{@{}c@{}}agreeing\\      \small{174; 4.5\%}\end{tabular}       & \begin{tabular}[c]{@{}c@{}}neutral\\      \small{176; 4.6\%}\end{tabular}         \\
\begin{tabular}[c]{@{}c@{}}confident\\      \small{18; 0.5\%}\end{tabular}       & \begin{tabular}[c]{@{}c@{}}confident\\      \small{20; 0.5\%}\end{tabular}       & \begin{tabular}[c]{@{}c@{}}wishing\\      \small{74; 1.9\%}\end{tabular}         & \begin{tabular}[c]{@{}c@{}}wishing\\      \small{98; 2.6\%}\end{tabular}          & \begin{tabular}[c]{@{}c@{}}consoling\\      \small{150; 3.9\%}\end{tabular}      & \begin{tabular}[c]{@{}c@{}}consoling\\      \small{170; 4.4\%}\end{tabular}       \\
\begin{tabular}[c]{@{}c@{}}suggesting\\      \small{10; 0.3\%}\end{tabular}      & \begin{tabular}[c]{@{}c@{}}suggesting\\      \small{10; 0.3\%}\end{tabular}      & \begin{tabular}[c]{@{}c@{}}joyful\\      \small{34; 0.9\%}\end{tabular}          & \begin{tabular}[c]{@{}c@{}}suggesting\\      \small{36; 0.9\%}\end{tabular}       & \begin{tabular}[c]{@{}c@{}}confident\\      \small{38; 1.0\%}\end{tabular}         & \begin{tabular}[c]{@{}c@{}}suggesting\\      \small{68; 1.8\%}\end{tabular}       \\
\begin{tabular}[c]{@{}c@{}}wishing\\      \small{8; 0.2\%}\end{tabular}          & \begin{tabular}[c]{@{}c@{}}wishing\\      \small{10; 0.3\%}\end{tabular}         & \begin{tabular}[c]{@{}c@{}}confident\\      \small{30; 0.8\%}\end{tabular}       & \begin{tabular}[c]{@{}c@{}}confident\\      \small{24; 0.6\%}\end{tabular}        & \begin{tabular}[c]{@{}c@{}}suggesting\\      \small{36; 0.9\%}\end{tabular}      & \begin{tabular}[c]{@{}c@{}}confident\\      \small{38; 1.0\%}\end{tabular}          \\ \bottomrule
\end{tabular}%
}
\caption{\label{tab:top-intents}
Top-10 most frequent emotion and intent labels across evaluated chatbots' responses per dialog turn. For each turn, the first column corresponds to counts in the original iEval dataset and the second one -- to counts in 
the logs generated during LLM-to-bot play.}
\end{table*}

We used iEval dataset for this experiment \cite{svikhnushina-etal-2022-ieval}. The dataset features human conversations with four empathetic chatbots collected in an online interactive manner. During the dataset curation process, each human was assigned an emotion label with the situation description taken from the EmpatheticDialogues dataset \cite{rashkin-etal-2019-towards} and asked to have a 6-turn conversation with each chatbot while playing a character in the assigned scenario. Overall, there are 480 situation descriptions in the dataset, which evenly cover two emotional polarities: positive and negative. As each chatbot participated in each scenario, there are in total of 1920 dialogs in the dataset. After conversing with the chatbots, human interlocutors provided their appraisals of chatbot listeners in each dialog, including five fine-grained listener qualities on a 5-point Likert scale: politeness, empathy, likability, repetitiveness, and making sense, and an overall dialog rating on a 3-point scale. All scores are provided on a dialog-level.

The four chatbot models used to curate the dataset were Blender \cite{roller-etal-2021-recipes}, MIME \cite{majumder-etal-2020-mime}, MEED and Plain \cite{xie-pu-2021-empathetic}. All of them are publicly available. We use these models in the same configurations for our experiment.

\subsubsection{LLM-to-Bot Play Results} \label{sec:llm-to-bot}

As the first step to validate our evaluation framework, we analyzed whether the LLM succeeds in mimicking human discourse following an assigned social role and whether approximating human speakers with the LLM causes any considerable changes in the chatbots' response patterns.

\begin{figure*}[!t]
    \centering
    \includegraphics[width=\linewidth]{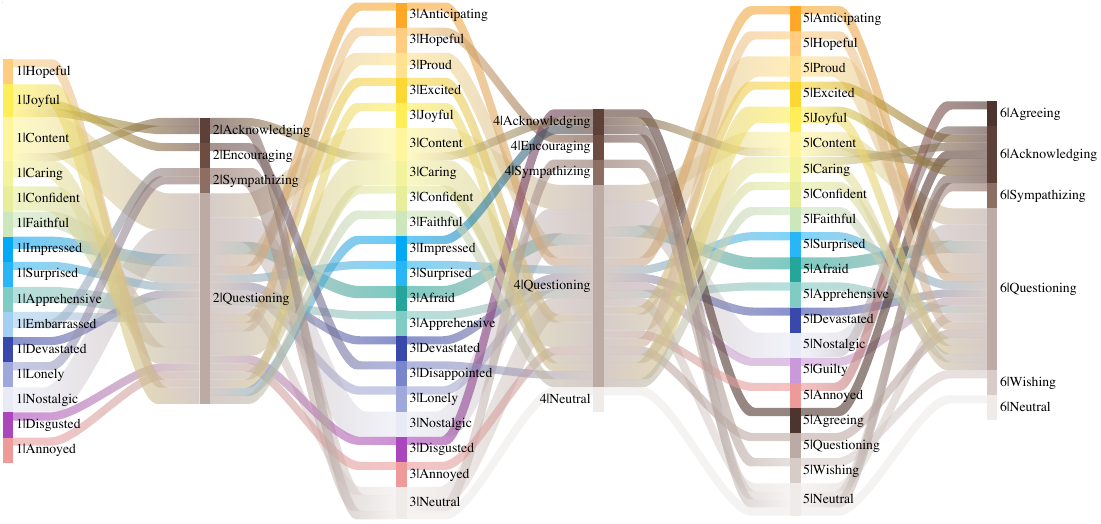}
    \caption{Sankey diagram showing discourse patterns in human-to-bots conversations originating from the iEval dataset.}
    \label{fig:h-b}
\end{figure*}

\begin{figure*}[!h]
    \centering
    \includegraphics[width=\linewidth]{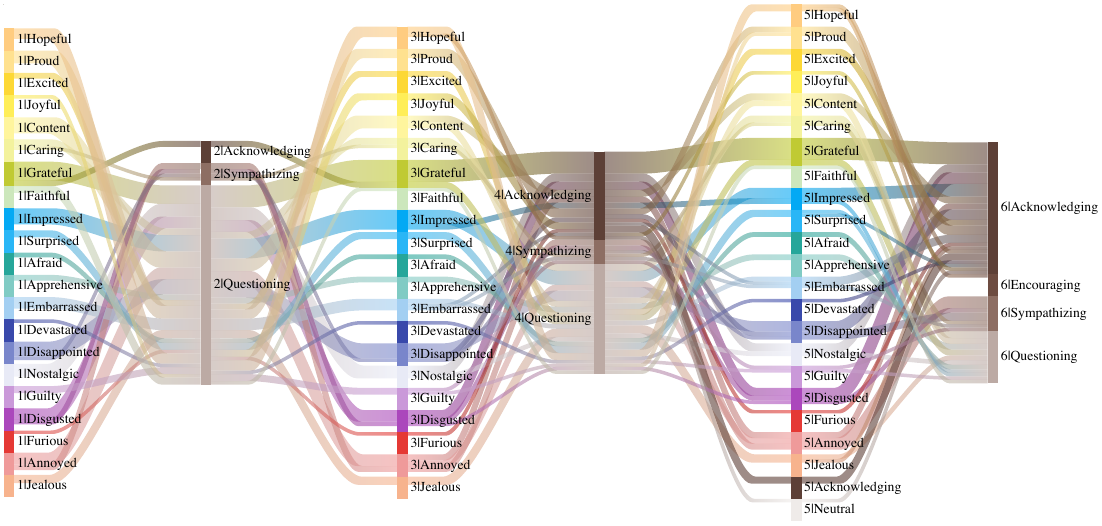}
    \caption{Sankey diagram showing discourse patterns in freshly generated LLM-to-bots conversations.}
    \label{fig:llm-b}
\end{figure*}

To generate LLM-to-bots conversations, we closely followed the procedure of iEval dataset curation. Specifically, we used emotion labels and situation descriptions from the dataset to create prompts for the LLM: \textit{I am a Speaker, feeling <emotion> because <situation>. I am sharing these emotions with a Listener, expecting empathy and understanding from them. I respond as a Speaker in a dialog.} The first LLM input was also taken from the iEval dataset. For each scenario, we collected LLM conversations with each of the four bots, letting them converse for 6 turns, i.e., 3 inputs from the LLM and 3 responses from the chatbot.

To examine the similarity of discourse patterns between human-to-bots and LLM-to-bots conversations, we started by annotating each dialog turn in two datasets with emotion and empathetic intent labels, using emotion/intent classifier developed by \citet{welivita-pu-2020-taxonomy} for EmpatheticDialogues dataset. As datasets in our experiment were grounded in situation descriptions taken from EmpatheticDialogues, the classifier was expected to generalize well to our data. 

\begin{figure*}[t]
    \centering
    \includegraphics[width=\linewidth]{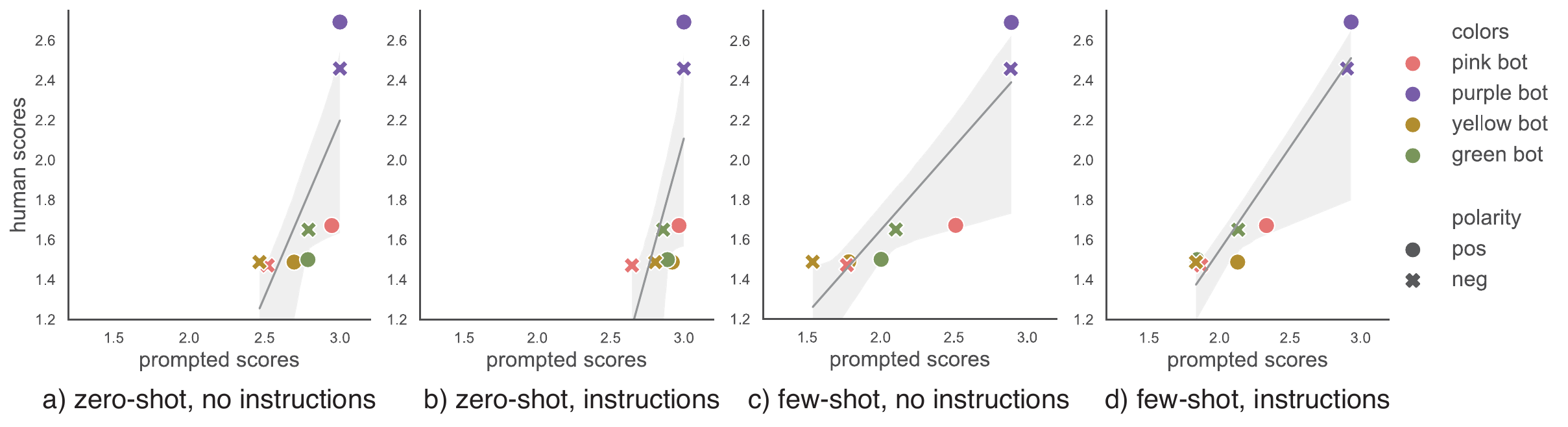}
    \caption{Scatter plots depicting the system-level correlation results. Human scores are based on the iEval dialog annotations, while prompted LLM scores are computed based on the generated dialogs.}
    \label{fig:corr}
\end{figure*}

Consequently, we visualized the most prominent discourse patterns\footnote{\textit{Pattern} implies an ordered sequence of emotion/intent labels expressed by speakers and listeners over the course of 6 dialog turns.} for two corpora in the form of Sankey diagrams, shown in Figures \ref{fig:h-b} and \ref{fig:llm-b}. The diagrams depict the flow connecting emotions expressed by the speakers and intents expressed by the listeners across dialog turns. Each odd step in the diagrams corresponds to human or LLM turns, while each even step summarizes intents and emotions in the responses of evaluated chatbots. To avoid clutter, we visualized patterns whose frequency exceeded a certain threshold.\footnote{We used a minimum frequency of 3 for the iEval dataset and a minimum frequency of 5 for the generated dataset.} From the visual inspection, it can be seen that the LLM emotion distribution over the course of the dialog (Figure \ref{fig:llm-b}) largely resembles one of the human interlocutors (Figure \ref{fig:h-b}). More importantly, sets of intents produced by empathetic chatbots are also very similar between the two figures, with \textit{Questioning}, \textit{Sympathizing}, and \textit{Acknowledging} being the most prominent ones. A quantitative comparison of the top 10 most prominent chatbots' intents and emotions across turns is shown in Table \ref{tab:top-intents}. Thus, our freshly generated interactive dataset with LLM-to-bot play was deemed to produce a reasonable approximation of human-to-bot conversations.

\subsubsection{Prompted Evaluation Results} \label{sec:prompted_eval}

Turning to the second step of our evaluation framework, we examined different types of prompting to produce scores for the generated LLM-to-bot dialogs. Specifically, two variables in the prompt design were considered.

\begin{table}[!b]
\centering
\begin{tabular}{l|cc}
\toprule
          & \multicolumn{1}{l}{No instructions}                       & \multicolumn{1}{l}{Instructions}                                    \\ \hline
Zero-shot & \begin{tabular}[c]{@{}c@{}}0.748\\ (p=0.033)\end{tabular} & \begin{tabular}[c]{@{}c@{}}0.651\\ (p=0.080)\end{tabular}           \\
Few-Shot  & \begin{tabular}[c]{@{}c@{}}0.892\\ (p=0.003)\end{tabular} & \begin{tabular}[c]{@{}c@{}}\textbf{0.954}\\ (p\textless{}0.001)\end{tabular} \\ \bottomrule
\end{tabular}%
\caption{\label{tab:ieval_corr}
System-level Pearson correlation for four possible prompt design manipulations, with the p-value in brackets.
}
\end{table}

First, we tried score generation in zero-shot and few-shot settings. For the few-shot setting, the number of demonstrations was fixed to the number of points in the ground truth human evaluation scale, with one representative example supplied for each score. Thus, for the iEval dataset, we used three demonstration dialogs corresponding to the three possible evaluation scores: \textit{Bad}, \textit{Okay}, and \textit{Good}. The examples were selected manually and are provided in Table \ref{fig:ieval_prompt_details} in Appendix \ref{sec:appendix_ieval}.

Second, we analyzed whether providing additional instructions helped the LLM evaluation performance. To write the instructions, we relied on the findings of \citet{svikhnushina-etal-2022-ieval}, which explained how chatbots' performance on various fine-grained dimensions translates into the overall score. As the authors emphasized the difference in humans' expectations of an empathetic listener in positive and negative conversational scenarios, we devised slightly different instructions to prompt the evaluation of these two emotional polarities. Specific formulations of the instructions are also provided in Table \ref{fig:ieval_prompt_details} in Appendix \ref{sec:appendix_ieval}.

To generate scores for each dialog, we prompted the LLM to complete the masked score, provided the log of the evaluated dialog. Depending on the configuration, few-shot demonstrations and/or instructions were prepended to the prompt. A template of the used prompt can be found in Figure \ref{fig:ieval_prompt_structure} in Appendix \ref{sec:appendix_ieval}. After obtaining dialog-level scores, we aggregated them to produce system-level ratings. One system was defined as a chatbot operating in one of the two emotional polarities. This decision is driven by the fact that based on human evaluation results in \cite{svikhnushina-etal-2022-ieval}, chatbots demonstrated statistically significant differences in their performance depending on the emotion. Thus, we considered eight systems for computing system-level correlations.

System-level correlations between human- and LLM-judgments for each of the four possible prompt design manipulations are reported in Table \ref{tab:ieval_corr}. Few-shot prompting with instructions results in the highest correlation of 0.954, which is further illustrated by the scatter plots in Figure \ref{fig:corr}. According to the plots, providing examples helps the LLM to calibrate the produced scores, eliminating the positivity bias, whereas instructions result in reduced variance.

\subsection{Generalizability to Different Domains}

In this section, we consider how prompted evaluation can generalize to different corpora and conversational settings. As the results above suggested that prompts combining instructions with examples perform best for evaluation, for the following experiment we searched for datasets that allowed formulating instructions for defining what properties correspond to good or bad overall appraisal ratings of the dialogs. Therefore, we selected two datasets that contained both fine-grained and overall ratings of the dialogs and used the information of the most relevant fine-grained dimensions to formulate instructions. We also considered only those datasets that contained multi-turn dialogs collected following the interactive process.

The selected datasets feature human-to-bot dialogs, with some dialog systems that are not publicly available. Moreover, these dialogs were collected in a generic manner, without the purpose to model any specific social behavior (e.g., as empathy in iEval). Due to these considerations, in the following experiments, we only studied the performance of the second step of our DEP framework, skipping the synthesis of new LLM-to-bots conversations. In a general case, when researchers have access to their evaluation targets, prompting LLMs to engage in a generic social interaction with the evaluated bots should be straightforward as we demonstrated in Section \ref{sec:llm-to-bot}.

\subsubsection{Datasets}
To study the generalizability of prompted evaluation, we used FED \cite{mehri-eskenazi-2020-unsupervised} and DSTC9 datasets \cite{gunasekara2020overview}. FED contains 124 open-domain dialogs of humans with humans and two chatbots (Meena and Mitsuku) that were originally released by \cite{adiwardana2020humanlike}. DSTC9 contains 2200 human-bot conversations from 11 chatbots. In both datasets, all dialogs are annotated with offline human appraisals of ten fine-grained dialog qualities and an overall impression rating that were curated following the same protocol described in \cite{mehri-eskenazi-2020-unsupervised}.

\subsubsection{Prompted Evaluation Results} \label{sec:prompted_eval_ext}
To construct a prompt for evaluating the chosen datasets, we selected five dialog examples covering five possible scores for overall dialog ratings, ranging from \textit{Very bad} to \textit{Very good}; they are provided in Table \ref{fig:fed_dialogs} in Appendix \ref{sec:appendix_fed}. To formulate the instructions, we used information from the original paper describing the relative importance of each fine-grained dialog quality for the overall impression. The specific formulation of the instruction is provided in Appendix \ref{sec:appendix_fed}.

The evaluation results with a comparison to existing best-performing evaluation metrics are provided in Table \ref{tab:ext_corr}. As the number of systems in the FED dataset is small, we only report dialog-level correlation. We also report Spearman correlation for this dataset for the purpose of comparison with the results in the original paper ($r=0.443$ ($p<0.05$)) \cite{mehri-eskenazi-2020-unsupervised}. Our prompted evaluation exceeds correlations of previous metrics by a considerable margin on both datasets and, thus, demonstrates the ability to generalize to new open-domain conversational settings.

\begin{table}[]
\centering
\resizebox{\linewidth}{!}{%
\begin{tabular}{lccc}
\hline
                                                                                     & FED                                                        & \multicolumn{2}{c}{DSTC9}                                                                                                 \\
                                                                                     & Dialog (S)                                           & Dialog (P)                                            & System (P)                                            \\ \hline
\multicolumn{1}{c}{\begin{tabular}[c]{@{}c@{}}Prev. best\\ (metric)\end{tabular}} & \begin{tabular}[c]{@{}c@{}}0.547\\ \shortcite{zhang-etal-2021-dynaeval} \end{tabular} & \begin{tabular}[c]{@{}c@{}}0.147\\ \shortcite{li-etal-2021-conversations} \end{tabular} & \begin{tabular}[c]{@{}c@{}}0.907\\ \shortcite{li-etal-2021-conversations} \end{tabular} \\
\multicolumn{1}{c}{DEP}                                                              & \textbf{0.655}                                                      & \textbf{0.274}                                                       & \textbf{0.980}                                                       \\ \hline
\end{tabular}%
}
\caption{\label{tab:ext_corr}
Results on FED and DSTC9 data. Previous best results are obtained from \cite{yeh-etal-2021-comprehensive}. Dialog and System indicate dialog- and system-level correlations, respectively, with P standing for Pearson and S for Spearman correlation. All values are statistically significant to $p<0.05$.
}
\end{table}

\section{Discussion}
Dialog system evaluation with prompting showed its usefulness both for generating new interactive exchanges with the evaluated systems and for judging their performance, therefore, allowing for a reasonable approximation of the online user evaluation pipeline.
We deem this approach particularly promising for the evaluation of social aspects of conversations. LLMs used for prompting suffer from occasional hallucinations, i.e., a tendency to make up factual information \cite{ouyang2022training}. It might be difficult to keep track of all specific factual items of information that come up in the interactively created dialog between two conversational models and search for ground truth references for each of them to construct objective metrics such as the model's accuracy or truthfulness \cite{lin-etal-2022-truthfulqa}. Whereas, prompting the LLM to establish a specific behavior and providing instructions about commonsense social norms appears more feasible once these instructions are established.

Drawing from the visualization of discourse patterns in our newly collected dataset of dialogs between the LLM and empathetic chatbots, we observed that the prompted LLM largely mirrors the conversational patterns of humans. However, there are also some differences. For example, in Figure \ref{fig:llm-b} there is an apparent sub-flow with a \textit{Grateful} emotion, increasingly displayed by the LLM. We believe the LLM might have developed an agreeable ``personality'' due to its training procedure based on Reinforcement Learning from Human Feedback, which optimized LLM's responses to satisfy human labelers. Differences in speakers' behavior led to the difference in the responses of the evaluated chatbots. While their most frequently produced intents are similar, their frequency distributions are statistically identical only for the second turn (first response of the evaluated chatbots) according to the permutation and chi-square tests. Future research can consider alternative prompting techniques to make the emotion/intent distribution of LLMs' and chatbots' responses even more balanced and representative. It might be beneficial to conduct additional experiments to compare original and generated dialogs, which can, for example, include testing the human ability to distinguish the dialogs created with the help of an LLM and dialogs with human speakers.

We conducted our experiments with only one LLM and explored the few-shot prompting scenarios with a fixed number of demonstrations. Future studies could explore the applicability of other LLMs for the DEP framework, as it has been already initiated by \cite{huynh2023understanding}. An area of particular interest would be to study the efficacy of the framework working with open-source LLMs, such as LLaMa \cite{touvron2023llama}. Additional investigation is necessary to analyze the capability of the framework to handle longer dialogs, which might be challenging to fit into a context window of an LLM.

We would also like to explore how DEP generalizes to evaluating other phenomena in social conversations, apart from generic open-domain interactions and empathetic dialogs. For example, further studies might focus on applying the framework to evaluate toxicity or humor in dialogs. However, this research direction requires the curation of appropriate calibration datasets.

Last but not least, evaluation artifacts produced by DEP may be used to assist designers of chatbots as they allow for both analyzing the synthesized logs and comparing quality ratings. These insights may be integrated into assistive chatbot design tools, such as \textit{iChatProfile} \cite{Han2021}, to offer a faster prototyping cycle due to the automatic generation of chat logs and richer insight about chatbot profiles due to additional rating information provided by the last step of DEP.

\section{Conclusion}
In this paper, we proposed DEP -- a framework for evaluating social chatbots using prompting. Our framework addresses the limitations of evaluation approaches using benchmark datasets in an offline setting. We describe how LLMs can be leveraged to synthesize realistic conversational logs with the evaluated chatbots in an online interactive manner. We further outline how the knowledge about the desired fine-grained qualities of a conversational partner can be translated into the prompting instructions to generate reliable overall scores for the collected dialogs. The proposed framework streamlines the evaluation process, making it highly efficient in terms of both time and cost, by removing the need for human involvement at every step. 
Our experiments demonstrated that the prompting-based evaluation results achieve a high correlation with human judgment, reaching an impressive Pearson $r = 0.95$ system-level correlation for the iEval dataset, which features dialogs with empathetic chatbots. We explain our vision of why this framework is well-suited for the evaluation of social phenomena in conversations and lay out future research directions. We also publicly release all freshly curated chat logs between the LLM and evaluated chatbots, as well as all additional annotations for the iEval, FED, and DSTC9 datasets created for this study.\footnote{\url{https://github.com/Sea94/dep}}

\section*{Acknowledgements}
This project has received funding from the Swiss National Science Foundation (Grant No.
200021\_184602). The authors also express gratitude to Mohamed Elasfoury for the invaluable help with conducting the generalizability experiments.

\bibliography{custom}
\bibliographystyle{acl_natbib}

\vfil
\clearpage

\appendix

\section{Prompt format for iEval}
\label{sec:appendix_ieval}

The template of a prompt used for producing scores for empathetic chatbots is provided in Figure \ref{fig:ieval_prompt_structure}. Depending on the prompting setting, either demonstrations, or instruction, or both were omitted from the prompt. For demonstrations, we used data in the same format as in the outlined box, but filling the mask score with the appropriate textual value. Dialogs used for demonstrations are included in Table \ref{fig:ieval_prompt_details}. If the instruction was used, we inserted the respective string in the prompt. The instructions that we used are also provided in Table \ref{fig:ieval_prompt_details}.

\begin{table}[!b]
\centering

\resizebox{\linewidth}{!}{%
\begin{tabular}{||l l||}
\hline
\multicolumn{2}{|p{\linewidth}|}{} \\ 
\multicolumn{2}{|p{\linewidth}|}{<\textit{demonstration \#1}>} \\ 
\multicolumn{2}{|p{\linewidth}|}{\textit{<demonstration \#2>}} \\ 
\multicolumn{2}{|p{\linewidth}|}{\textit{<demonstration \#3>}} \\ 
\multicolumn{2}{|p{\linewidth}|}{} \\ 
\hhline{||--||}
\multicolumn{2}{||p{\linewidth}||}{I am a Speaker, feeling <\textit{emotion}> because <\textit{situation}>. I shared these emotions with a Listener in a dialog, expecting empathy and understanding from them. Our dialog went as follows.} \\[10ex]
\multicolumn{2}{||p{\linewidth}||}{Speaker:  \textit{<LLM's input \#1>}}              \\
\multicolumn{2}{||p{\linewidth}||}{Listener:  \textit{<Bot's response \#1>}}                    \\
\multicolumn{2}{||p{\linewidth}||}{Speaker:  \textit{<LLM's input \#2>}}              \\
\multicolumn{2}{||p{\linewidth}||}{Listener:  \textit{<Bot's response \#2>}}                    \\   
\multicolumn{2}{||p{\linewidth}||}{Speaker:  \textit{<LLM's input \#3>}}                    \\
\multicolumn{2}{||p{\linewidth}||}{Listener:  \textit{<Bot's response \#3>}}                    \\[1.2ex]
\multicolumn{2}{||p{\linewidth}||}{<\textit{Instruction}> I would rate the Listener in my dialog as \_\_\_, choosing from Bad, Okay, and Good options.} \\ \hhline{||--||}
\multicolumn{2}{|p{.8\linewidth}|}{} \\ \hline
\end{tabular}
}
\captionof{figure}{Prompt template for evaluating empathetic listeners in freshly collected
dialogs of empathetic models from iEval with the LLM.} \label{fig:ieval_prompt_structure}
\end{table}

\begin{table}[!h]
\centering
\resizebox{!}{0.49\textheight}{%
\begin{tabular}{l}
\toprule
\begin{tabular}[c]{@{}p{\linewidth}@{}}
\small{\textit{\textbf{Speaker:}} Hi!}\\ 
\small{\textit{\textbf{Listener:}} Hi there.}\\
\small{\textit{\textbf{Speaker:}} I want a recommendation for a holiday destination} \\
\small{\textit{\textbf{Listener:}} Have you tried asking your friends what they like?}\\
\small{\textit{\textbf{Speaker:}} I have, but I'm looking for your point of view}\\ 
\small{\textit{\textbf{Listener:}} What was the reply? Have you tried looking in a newspaper article?}\\
\small{\textit{\textbf{Speaker:}} Sorry? I said I want your point of view}\\ 
\small{\textit{\textbf{Listener:}} It's OK. After all, you are only human. My opinion is of no consequence.}\\
\small{\textit{\textbf{Speaker:}} Yours is the opinion I want}\end{tabular}       \\ \hline

\begin{tabular}[c]{@{}p{\linewidth}@{}}
\small{\textit{\textbf{Speaker:}} Hi!}\\ 
\small{\textit{\textbf{Listener:}} Hi there.}\\
\small{\textit{\textbf{Speaker:}} I'm trying to figure out what to make for this weekend's party. Any suggestions?} \\
\small{\textit{\textbf{Listener:}} Don't think too hard. I'm sure I can smell sawdust. First you must download me to your personal computer.}\\
\small{\textit{\textbf{Speaker:}} I must do what the what now?}\\ 
\small{\textit{\textbf{Listener:}} Right now?Why do you have to do it?}\\
\small{\textit{\textbf{Speaker:}} Hey, your spacing is off.}\\ 
\small{\textit{\textbf{Listener:}} You've got my full attention. off was not my intention.}\\ 
\small{\textit{\textbf{Speaker:}} And your capitalization!}\end{tabular}       \\ \hline

\begin{tabular}[c]{@{}p{\linewidth}@{}}
\small{\textit{\textbf{Speaker:}} Hi!}\\ 
\small{\textit{\textbf{Listener:}} Hi! How are you today?}\\
\small{\textit{\textbf{Speaker:}} What's laser tag?} \\
\small{\textit{\textbf{Listener:}} Like paintball, but with lasers!}\\
\small{\textit{\textbf{Speaker:}} lol good description}\\ 
\small{\textit{\textbf{Listener:}} Do you know what paintball is?}\\
\small{\textit{\textbf{Speaker:}} yeah I played it before}\\ 
\small{\textit{\textbf{Listener:}} Cool! What did you think?}\\
\small{\textit{\textbf{Speaker:}} It's somewhat exciting, but very tiring :)}\\ 
\small{\textit{\textbf{Listener:}} That is very true. What is your favorite color?}\\
\small{\textit{\textbf{Speaker:}} I like red}\end{tabular}       \\ \hline

\begin{tabular}[c]{@{}p{\linewidth}@{}}
\small{\textit{\textbf{Speaker:}} Hi!}\\ 
\small{\textit{\textbf{Listener:}} What is your favorite holiday?}\\
\small{\textit{\textbf{Speaker:}} one where I get to meet lots of different people.} \\
\small{\textit{\textbf{Listener:}} What was the most number of people you have ever met during a holiday?}\\
\small{\textit{\textbf{Speaker:}} Hard to keep a count. Maybe 25.}\\ 
\small{\textit{\textbf{Listener:}} Which holiday was that?}\\
\small{\textit{\textbf{Speaker:}} I think it was Australia}\\ 
\small{\textit{\textbf{Listener:}} Do you still talk to the people you met?}\\
\small{\textit{\textbf{Speaker:}} Not really. The interactions are usually short-lived but it's fascinating to learn where people are coming from and what matters to them}\end{tabular}       \\ \hline

\begin{tabular}[c]{@{}p{\linewidth}@{}}
\small{\textit{\textbf{Speaker:}} Hi!}\\ 
\small{\textit{\textbf{Listener:}} Hi! How's it going?}\\
\small{\textit{\textbf{Speaker:}} Good! How are you?} \\
\small{\textit{\textbf{Listener:}} I'm well, thanks! How was your day?}\\
\small{\textit{\textbf{Speaker:}} My day was fine, I just went to work today. How was your day?}\\ 
\small{\textit{\textbf{Listener:}} My day was fine. I've been procrastinating on finishing my homework, but it's due in a few weeks, so I'll get it done eventually. I've watched a bunch of anime today. Where do you work?}\\
\small{\textit{\textbf{Speaker:}} I work at a large tech company}\\ 
\small{\textit{\textbf{Listener:}} Cool! What do you do for the company?}\\
\small{\textit{\textbf{Speaker:}} I work on machine learning research}\end{tabular}       \\ \bottomrule
\end{tabular}
}
\caption{\label{fig:fed_dialogs}
Demonstrations used for FED prompts, ranging from Very bad (top) to Very good (bottom).
}
\end{table}

\begin{table*}[!ht]
\centering
\resizebox{\linewidth}{!}{%
\begin{tabular}{l|l|l}
\toprule
& \multicolumn{1}{c|}{\textbf{Positive}}                                                                                                                                                                                                                                                                                                                                                                                                & \multicolumn{1}{c}{\textbf{Negative}}                                                                                                                                                                                                                                                                                                                                                                                                                                            \\ \hline
Instructions & \multicolumn{1}{p{0.4\linewidth}|}{\small{In positive contexts, like this one, good empathetic listeners always respond politely and demonstrate attention. More importantly, they try to amplify speaker's positive emotion by asking follow-up questions and sharing their appraisal of the situation. On the contrary, bad empathetic listeners repeat themselves too much and don't follow the context.}}   & \multicolumn{1}{p{0.4\linewidth}}{\small{In negative contexts, like this one, good empathetic listeners always respond politely and demonstrate attention. More importantly, they try to clarify the context and the consequences for the speaker and alleviate speaker's negative emotion by sympathizing and suggesting solutions. On the contrary, bad empathetic listeners ignore speaker's emotion, ask inappropriate questions, repeat themselves too much and focus on self instead of the speaker.}} \\

\midrule
Bad & \begin{tabular}[c]{@{}p{.4\linewidth}@{}}\small{\textit{\textbf{Speaker:}} I had a pretty large loan, with a bit of a high interest rate, and a high monthly payment. My mother decided to pay it off for me, out of the blue!}\\
\small{\textit{\textbf{Listener:}} that is a shame. how long have you had to do? that sounds like you have a good relationship with your mom?}\\ 
\small{\textit{\textbf{Speaker:}} I have been paying off this loan for several months. I have such a good relationship with my mother that she relieved me of this debt much to my surprise.}\\ 
\small{\textit{\textbf{Listener:}} that sounds like a great thing to hear}\\ 
\small{\textit{\textbf{Speaker:}} Yes, I am very happy to not have to make monthly payments to pay off this high interest rate loan anymore.}\\ \small{\textit{\textbf{Listener:}} that is a good feeling. i am sure you will get the job!}\end{tabular}                         & \begin{tabular}[c]{@{}p{.4\linewidth}@{}} \small{\textit{\textbf{Speaker:}} I was one percent off from passing my math test, I was devastated.}\\ 
\small{\textit{\textbf{Listener:}} i'm sorry to hear that. were you able to get a better grade on the test?}\\ \small{\textit{\textbf{Speaker:}} No, I am just upset.}\\ \small{\textit{\textbf{Listener:}} i think i am going to go back to school. i am not sure what i will do.}\\ \small{\textit{\textbf{Speaker:}} Make sure to study.}\\ \small{\textit{\textbf{Listener:}} i am sure you will do great. i hope you get a good grade on your test. good luck!}\end{tabular} \\ \hline

Okay & \begin{tabular}[c]{@{}p{.4\linewidth}@{}} \small{\textit{\textbf{Speaker:}} My son drove down and spent the whole weekend helping me move.}\\ 
\small{\textit{\textbf{Listener:}} That's great! How old is he?}\\
\small{\textit{\textbf{Speaker:}} He's going to be turning 30 this year. He's such a sweet son.} \\
\small{\textit{\textbf{Listener:}} That's awesome. I'm happy for him.}\\
\small{\textit{\textbf{Speaker:}} Thank you. Moving is such a pain, it's always  nice to have help.}\\ 
\small{\textit{\textbf{Listener:}} hat's great. I'm happy for you.}\end{tabular} & \begin{tabular}[c]{@{}p{.4\linewidth}@{}} \small{\textit{\textbf{Speaker:}} I was recently on a long international flight and we hit some really bad turbulence.}\\ 
\small{\textit{\textbf{Listener:}} Oh no, what happened?}\\ 
\small{\textit{\textbf{Speaker:}} The flight attendants weren't able to do much for us, unfortunately.}\\
\small{\textit{\textbf{Listener:}} Oh no, what happened?}\\
\small{\textit{\textbf{Speaker:}} You don't have to repeat yourself. We had turbulence on the flight and the attendants didn't help us.}\\
\small{\textit{\textbf{Listener:}} That's awful. I'm glad you were okay.}\end{tabular}       \\ \hline

Good & \begin{tabular}[c]{@{}p{.4\linewidth}@{}} \small{\textit{\textbf{Speaker:}} I am going on a vacation this Thursday! I am very excited!}\\ 
\small{\textit{\textbf{Listener:}} that's awesome! where are you going? i'm sure you'll have a great time!}\\
\small{\textit{\textbf{Speaker:}} Thanks, we're going to see the Grand Canyon.} \\
\small{\textit{\textbf{Listener:}} that sounds like a lot of fun! i've never been there, but i hear it's beautiful.}\\
\small{\textit{\textbf{Speaker:}} Me too!}\\ 
\small{\textit{\textbf{Listener:}} i'd love to go on a cruise one day. i hope you have a wonderful time!}\end{tabular} & \begin{tabular}[c]{@{}p{.4\linewidth}@{}} \small{\textit{\textbf{Speaker:}} I was out walking by the lake over the weekend and there shore was just covered in dead rotting fish.}\\ 
\small{\textit{\textbf{Listener:}} Oh no! Are you ok?}\\ 
\small{\textit{\textbf{Speaker:}} Yes, I'm okay. It was just weird to see so many dead fish}\\
\small{\textit{\textbf{Listener:}} I bet that was scary.}\\
\small{\textit{\textbf{Speaker:}} Yes, I would definitely not want to encounter that experience again.}\\
\small{\textit{\textbf{Listener:}} That sounds like a scary experience. I'm glad you are ok.}\end{tabular}       \\ \bottomrule
\end{tabular}
}
\caption{\label{fig:ieval_prompt_details}
Instructions and demonstration used for prompts for evaluating empathetic listeners in freshly collected dialogs of empathetic models from iEval with the LLM. Demonstrations and their appraisals are manually selected from the iEval dataset. Inputs from ``Positive'' column were used for dialogs conditioned on positive emotion label and inputs from ``Negative'' column -- for dialogs conditioned on negative emotion label.
}
\end{table*}

\section{Prompt format for FED}
\label{sec:appendix_fed}

While working with FED dataset, we used a similar template as shown in Figure \ref{fig:ieval_prompt_structure}. In a few-shot setting, we included five demonstrations instead of three, keeping one demonstration per possible rating value. The dialogs used for demonstrations are provided in Table \ref{fig:fed_dialogs}.

The instruction used for FED dataset was we following: \textit{In such open-ended dialogs, good listeners demonstrate coherence and maintain a good conversation flow, 
they display a likeable personality and understanding of the speaker. On the contrary, 
bad listeners don’t follow the context and don’t show much interest in the conversation.}

\end{document}